%% file: root.tex
\documentclass[letterpaper, 10 pt, conference]{ieeeconf}

\IEEEoverridecommandlockouts                              
                                                         
\usepackage{xcolor}
\usepackage{cite}
\usepackage{amsmath,amssymb,amsfonts}
\usepackage{algorithmic}
\usepackage{algorithm}
\usepackage{graphicx}
\usepackage{subfig}
\usepackage{textcomp}
\usepackage{verbatim}

\graphicspath{{pics/}}

\input{./include/pub_tools.tex}

\input{./include/glossaries.tex}

\title{\LARGE \bf
Gaussian Process-based Stochastic Model Predictive Control for Overtaking in Autonomous Racing
}

\author{T. Br\"udigam$^{1}$, A. Capone$^{2}$, S. Hirche$^{2}$, D. Wollherr$^{1}$, and M. Leibold$^{1}$
\thanks{$^{1}$The authors are with the Chair of Automatic Control Engineering at the Technical University of Munich, Arcisstrasse 21, 80333 Munich, Germany.
{\tt\small \{tim.bruedigam; dw; marion.leibold\}@tum.de}} 
\thanks{$^{2}$The authors are with the Chair of Information-oriented Control at the Technical University of Munich, Barer Strasse 21, 80333 Munich, Germany.
        {\tt\small \{alexandre.capone; hirche\}@tum.de}}%
}

\begin{document}

\maketitle
\thispagestyle{empty}
\pagestyle{empty}

\begin{abstract}
A fundamental aspect of racing is overtaking other race cars. Whereas previous research on autonomous racing has majorly focused on lap-time optimization, here, we propose a method to plan overtaking maneuvers in autonomous racing. A Gaussian process is used to learn the behavior of the leading vehicle. Based on the outputs of the Gaussian process, a stochastic Model Predictive Control algorithm plans optimistic trajectories, such that the controlled autonomous race car is able to overtake the leading vehicle. The proposed method is tested in a simple simulation scenario. 
\end{abstract}

\input{./chapters/introduction.tex}
\input{./chapters/problem.tex}
\input{./chapters/method.tex}
\input{./chapters/results.tex}
\input{./chapters/discussion.tex}

\input{./chapters/conclusion.tex}

\section*{Acknowledgement}

We thank Lukas Fichtner for valuable discussions.

\bibliography{./references/Dissertation_bib}
\bibliographystyle{unsrt}

\end{document}

%% file: include/pub_tools.tex
\usepackage{mathtools}

\usepackage[acronym, style=alttree, toc=true, shortcuts, xindy, nomain, nonumberlist]{glossaries}

\usepackage{cleveref}

\usepackage{hyphenat}
\usepackage{bm}

\usepackage{siunitx}  

\usepackage{booktabs}

\newcommand{\gvec}[1]{\boldsymbol{#1}}
\newcommand{\Mat}[1]{\mathbf{\MakeUppercase{#1}}}

\newcommand{\norm}[1]{\left\lVert#1\right\rVert} 
\newcommand{\set}[1]{\mathcal{#1}} 

\newcommand{\rv}[1]{\mathrm{#1}} 
\newcommand{\lr}[1]{\left(#1\right)} 

\newcommand{\T}[0]{T}

\DeclarePairedDelimiterXPP\onenorm[1]{}\lVert\rVert{_1}{\ifblank{#1}{\:\cdot\:}{#1}} 
\DeclarePairedDelimiterXPP\twonorm[1]{}\lVert\rVert{_2}{\ifblank{#1}{\:\cdot\:}{#1}} 

\newglossary{symbols}{sym}{sbl}{List of Symbols}
\newglossary{notation}{not}{nt}{Notation}

\glsaddkey{dimension}{\glsentrytext{\glslabel}}{\glsentryd}{\GLsentryd}{\glsd}{\Glsd}{\GLSd}
\glsaddkey{body}{\glsentrytext{\glslabel}}{\glsentrybody}{\GLsentrybody}{\glsbody}{\Glsbody}{\GLSbody}

%% file: include/glossaries.tex
\newacronym{ocp}{OCP}{optimal control problem}
\newacronym{MPC}{MPC}{Model Predictive Control}
\newacronym{MPCa}{MPC algorithm}{MPC algorithm}
\newacronym{RMPC}{RMPC}{Robust Model Predictive Control}
\newacronym{SMPC}{SMPC}{Stochastic Model Predictive Control}
\newacronym{SCMPC}{SCMPC}{Scenario Stochastic Model Predictive Control}
\newacronym{MILP}{MILP}{Mixed Integer Linear Program}
\newacronym{PIT}{PIT}{Pointwise\hyp{}In\hyp{}Time}
\newacronym{POMDP}{POMDP}{Partially Observable Markov Decision Process}
\newacronym{MDP}{MDP}{Markov Decision Process}
\newacronym{KKT}{KKT}{Karush\hyp{}Kuhn\hyp{}Tucker}
\newacronym{SMPCFT}{SMPC+FTP}{\textit{Stochastic Model Predictive Control + fail-safe trajectory planning}}
\newacronym{FT}{FTP}{fail-safe trajectory planning}
\newacronym{tv}{TV}{target vehicle}
\newacronym{ev}{EV}{ego vehicle}

\newglossaryentry{chiUk}{type=symbols,
	sort={control},
	name={\ensuremath{\boldsymbol{\chi}_h^{\bm{U}_h}}},
	description={?}
}
\newglossaryentry{chiX}{type=symbols,
	sort={control},
	name={\ensuremath{\boldsymbol{\chi}}},
	description={?}
}
\newglossaryentry{xik}{type=symbols,
	sort={control},
	name={\ensuremath{\bm{\xi}_k}},
	description={?}
}
\newglossaryentry{xi0}{type=symbols,
	sort={control},
	name={\ensuremath{\bm{\xi}_0}},
	description={?}
}
\newglossaryentry{sk}{type=symbols,
	sort={control},
	name={\ensuremath{s_k}},
	description={?}
}
\newglossaryentry{phik}{type=symbols,
	sort={control},
	name={\ensuremath{\phi_k}},
	description={?}
}
\newglossaryentry{dk}{type=symbols,
	sort={control},
	name={\ensuremath{d_k}},
	description={?}
}
\newglossaryentry{vk}{type=symbols,
	sort={control},
	name={\ensuremath{v_k}},
	description={?}
}
\newglossaryentry{uk}{type=symbols,
	sort={control},
	name={\ensuremath{\bm{u}_k}},
	description={?}
}
\newglossaryentry{ak}{type=symbols,
	sort={control},
	name={\ensuremath{a_k}},
	description={?}
}
\newglossaryentry{deltak}{type=symbols,
	sort={control},
	name={\ensuremath{\delta_k}},
	description={?}
}

\newglossaryentry{xitv}{type=symbols,
	sort={control},
	name={\ensuremath{\bm{\xi}^{\text{TV}}_k}},
	description={?}
}
\newglossaryentry{Rtv}{type=symbols,
	sort={control},
	name={\ensuremath{\mathcal{R}^{\text{TV}}_k}},
	description={?}
}
\newglossaryentry{bRtv}{type=symbols,
	sort={control},
	name={\ensuremath{\overline{\mathcal{R}}^{\text{TV}}_k}},
	description={?}
}
\newglossaryentry{xitv0}{type=symbols,
	sort={control},
	name={\ensuremath{\bm{\xi}^{\text{TV}}_0}},
	description={?}
}
\newglossaryentry{dxEVTV}{type=symbols,
	sort={control},
	name={\ensuremath{ \Delta x^\text{EV,TV}_0 }},
	description={?}
}

\newglossaryentry{hxitv0}{type=symbols,
	sort={control},
	name={\ensuremath{\bm{\hat{\xi}}^{\text{TV}}_0}},
	description={?}
}
\newglossaryentry{hxitv}{type=symbols,
	sort={control},
	name={\ensuremath{\bm{\hat{\xi}}^{\text{TV}}}},
	description={?}
}
\newglossaryentry{xitvref}{type=symbols,
	sort={control},
	name={\ensuremath{\bm{\xi}^{\text{TV}}_{\text{ref},k}}},
	description={?}
}
\newglossaryentry{utv}{type=symbols,
	sort={control},
	name={\ensuremath{\bm{u}^{\text{TV}}_k}},
	description={?}
}
\newglossaryentry{tutv}{type=symbols,
	sort={control},
	name={\ensuremath{\bm{\tilde{u}}^{\text{TV}}_k}},
	description={?}
}
\newglossaryentry{wtv}{type=symbols,
	sort={control},
	name={\ensuremath{\bm{w}^{\text{TV}}_k}},
	description={?}
}
\newglossaryentry{wsens}{type=symbols,
	sort={control},
	name={\ensuremath{\bm{w}^{\text{sens}}_0}},
	description={?}
}
\newglossaryentry{Wsens}{type=symbols,
	sort={control},
	name={\ensuremath{\mathcal{W}^{\text{sens}}}},
	description={?}
}
\newglossaryentry{xtv}{type=symbols,
	sort={control},
	name={\ensuremath{x^{\text{TV}}_k}},
	description={?}
}
\newglossaryentry{ytv}{type=symbols,
	sort={control},
	name={\ensuremath{y^{\text{TV}}_k}},
	description={?}
}
\newglossaryentry{vxtv}{type=symbols,
	sort={control},
	name={\ensuremath{v_{x,k}^{\text{TV}}}},
	description={?}
}
\newglossaryentry{vytv}{type=symbols,
	sort={control},
	name={\ensuremath{v_{y,k}^{\text{TV}}}},
	description={?}
}
\newglossaryentry{xtv0}{type=symbols,
	sort={control},
	name={\ensuremath{x^{\text{TV}}_0}},
	description={?}
}
\newglossaryentry{ytv0}{type=symbols,
	sort={control},
	name={\ensuremath{y^{\text{TV}}_0}},
	description={?}
}
\newglossaryentry{vxtv0}{type=symbols,
	sort={control},
	name={\ensuremath{v_{x,0}^{\text{TV}}}},
	description={?}
}
\newglossaryentry{vytv0}{type=symbols,
	sort={control},
	name={\ensuremath{v_{y,0}^{\text{TV}}}},
	description={?}
}
\newglossaryentry{evlane}{type=symbols,
	sort={control},
	name={\ensuremath{y_{\text{lane},0}^\text{EV}}},
	description={?}
}
\newglossaryentry{tvlane}{type=symbols,
	sort={control},
	name={\ensuremath{y_{\text{lane},0}^\text{TV}}},
	description={?}
}
\newglossaryentry{dclose}{type=symbols,
	sort={control},
	name={\ensuremath{r_\text{close}}},
	description={?}
}
\newglossaryentry{dcloseft}{type=symbols,
	sort={control},
	name={\ensuremath{r^\text{FTP}_\text{close}}},
	description={?}
}
\newglossaryentry{msafe}{type=symbols,
	sort={control},
	name={\ensuremath{\varepsilon_\text{safe}}},
	description={?}
}
\newglossaryentry{fclose}{type=symbols,
	sort={control},
	name={\ensuremath{r_\text{close}}},
	description={?}
}
\newglossaryentry{ffclose}{type=symbols,
	sort={control},
	name={\ensuremath{f_\text{close}}},
	description={?}
}
\newglossaryentry{fcloseft}{type=symbols,
	sort={control},
	name={\ensuremath{r^\text{FTP}_\text{close}}},
	description={?}
}
\newglossaryentry{ffcloseft}{type=symbols,
	sort={control},
	name={\ensuremath{f^\text{FTP}_\text{close}}},
	description={?}
}

\newglossaryentry{dt}{type=symbols,
	sort={control},
	name={\ensuremath{T}}, 
	description={?}
}

\newglossaryentry{wcov}{type=symbols,
	sort={control},
	name={\ensuremath{\Sigma^{\text{TV}}_{\bm{w}}}},
	description={?}
}
\newglossaryentry{wcov_t}{type=symbols,
	sort={control},
	name={\ensuremath{\tilde{\Sigma}^{\text{TV}}_{\bm{w}}}},
	description={?}
}
\newglossaryentry{wsenscov}{type=symbols,
	sort={control},
	name={\ensuremath{\Sigma^{\text{sens}}}},
	description={?}
}

\newglossaryentry{Xilane}{type=symbols,
	sort={control},
	name={\ensuremath{\mathcal{D}^{\text{lane}}}},
	description={?}
}

\newglossaryentry{UU}{type=symbols,
	sort={control},
	name={\ensuremath{\mathcal{U}}},
	description={?}
}
\newglossaryentry{XX}{type=symbols,
	sort={control},
	name={\ensuremath{\bm{\Xi}}},
	description={?}
}

\newglossaryentry{XXf}{type=symbols,
	sort={control},
	name={\ensuremath{\bm{\Xi}_\textnormal{f}}},
	description={?}
}
\newglossaryentry{XXsafe}{type=symbols,
	sort={control},
	name={\ensuremath{\bm{\Xi}_\textnormal{safe}}},
	description={?}
}

\newglossaryentry{Usafe}{type=symbols,
	sort={control},
	name={\ensuremath{\bm{U}_\textnormal{safe}}},
	description={?}
}
\newglossaryentry{Usafeh}{type=symbols,
	sort={control},
	name={\ensuremath{\bm{U}_{\textnormal{safe},h}}},
	description={?}
}
\newglossaryentry{Usmpc}{type=symbols,
	sort={control},
	name={\ensuremath{\bm{U}_\text{SMPC}}},
	description={?}
}
\newglossaryentry{Uft}{type=symbols,
	sort={control},
	name={\ensuremath{\bm{U}_\text{FTP}}},
	description={?}
}
\newglossaryentry{Uft2}{type=symbols,
	sort={control},
	name={\ensuremath{\bm{U}'
	_\text{FTP}}},
	description={?}
}
\newglossaryentry{Ubrake}{type=symbols,
	sort={control},
	name={\ensuremath{\bm{U}_\text{brake}}},
	description={?}
}

\newglossaryentry{Nsmpc}{type=symbols,
	sort={control},
	name={\ensuremath{N_\text{SMPC}}},
	description={?}
}

\newglossaryentry{Nft}{type=symbols,
	sort={control},
	name={\ensuremath{N_\text{FTP}}},
	description={?}
}

	\newglossaryentry{state}{type=symbols,
		sort={control},
		name={\ensuremath{\gvec{\xi}}},
		description={State vector}
	}
	
	\newglossaryentry{admissible_states}{type=symbols,
		sort={control},
		name={\ensuremath{\set{\Xi}_k}},
		description={Set of admissible states at time $k$}
	}
	
	\newglossaryentry{position_vector}{type=symbols,
		sort={control},
		name={\ensuremath{\gvec{\zeta}}},
		description={Position vector}
	}
	
	\newglossaryentry{rectangle_vertex}{type=symbols,
		sort={control},
		name={\ensuremath{\gvec{\zeta}^{\text{vert}}}},
		description={Vertex point of the safety rectangle on which the linear constraint is based}
	}
	
	\newglossaryentry{position}{type=symbols,
		sort={control},
		name={\ensuremath{x,y}},
		description={Longitudinal and lateral position}
	}
	
	\newglossaryentry{velo}{type=symbols,
		sort={control},
		name={\ensuremath{v_x, v_y}},
		description={Longitudinal and lateral velocity}
	}
	
	\newglossaryentry{control_input}{type=symbols,
		sort={control},
		name={\ensuremath{\vec{u}}},
		description={Control input vector}
	}
	
	\newglossaryentry{delta_control_input}{type=symbols,
		sort={control},
		name={\ensuremath{\triangle \vec{u}}},
		description={Rate of the control input vector \ensuremath{\vec{u}}}
	}
	
    \newglossaryentry{stacked_inputs}{type=symbols,
		sort={control},
		name={\ensuremath{U}},
		description={Sequence of input vectors}
	}
	
	\newglossaryentry{admissible_inputs}{type=symbols,
		sort={control},
		name={\ensuremath{\mathcal{U}_k}},
		description={Set of admissible inputs at time $k$}
	}
	
	\newglossaryentry{disturbance_Sens}{type=symbols,
		sort={control},
		name={\ensuremath{\vec{w}^{\text{Sens}}}},
		description={Disturbance vector corrupting the TV state measurement}
	}
	
	\newglossaryentry{disturbance_TV}{type=symbols,
		sort={control},
		name={\ensuremath{\vec{w}^{\text{TV}}}},
		description={Disturbance vector corrupting the TV inputs}
	}
	
	\newglossaryentry{disturbance_set}{type=symbols,
		sort={control},
		name={\ensuremath{\set{W}^{\text{Sens}}}},
		description={Compact, bounded set of measurement disturbances}
	}
	
	\newglossaryentry{Sens_covariance}{type=symbols,
		sort={control},
		name={\ensuremath{\set{\Sigma}^{\text{Sens}}}},
		description={Covariance matrix of the TV state measurement disturbance}
	}

	\newglossaryentry{TV_covariance}{type=symbols,
		sort={control},
		name={\ensuremath{\set{\Sigma}^{\text{TV}}_w}},
		description={Covariance matrix of the TV disturbance}
	}

	\newglossaryentry{sampling_time}{type=symbols,
		sort={control},
		name={\ensuremath{\triangle t}},
		description={Sampling time}
	}

	\newglossaryentry{vehicle_dim}{type=symbols,
		sort={control},
		name={\ensuremath{l_x, l_y}},
		description={Vehicle dimensions}
	}
	
	\newglossaryentry{lanewidth}{type=symbols,
		sort={control},
		name={\ensuremath{l_{\text{lane}}}},
		description={Lane width}
	}
	
	\newglossaryentry{lane_nr}{type=symbols,
		sort={control},
		name={\ensuremath{n_{\text{lane}}}},
		description={Number of lanes}
	}

	\newglossaryentry{rectangle_dim}{type=symbols,
		sort={control},
		name={\ensuremath{a,b}},
		description={Minimum safety rectangle dimensions}
	}
	
	\newglossaryentry{rectangle_length}{type=symbols,
		sort={control},
		name={\ensuremath{\tilde{a}}},
		description={Approximated length of the safety rectangle}
	}
	
	\newglossaryentry{delta_rectangle_dim}{type=symbols,
		sort={control},
		name={\ensuremath{\triangle a}},
		description={Velocity dependent longitudinal safety distance}
	}
	
	\newglossaryentry{delta_v_squared}{type=symbols,
		sort={control},
		name={\ensuremath{\triangle v^2}},
		description={Difference between the quadratic velocities of the EV and TV}
	}

    \newglossaryentry{approximation_c}{type=symbols,
		sort={control},
		name={\ensuremath{c}},
		description={Approximation parameter for the velocity dependent safety distance}
	}
	
	\newglossaryentry{steepness_parameter}{type=symbols,
		sort={control},
		name={\ensuremath{c_{\text{sd}}}},
		description={Parameter defining the steepness of the linear constraints}
	}
	
	\newglossaryentry{constraint_identification}{type=symbols,
		sort={control},
		name={\ensuremath{c_{\text{sc}}}},
		description={Parameter for the choice between regular and special case constraints}
	}
	
	\newglossaryentry{constraint}{type=symbols,
		sort={control},
		name={\ensuremath{d_k}},
		description={State constraint at time step $k$}
	}

    \newglossaryentry{pred_horizon}{type=symbols,
		sort={control},
		name={\ensuremath{N}},
		description={Length of the prediction horizon}
	}

    \newglossaryentry{prediction_error}{type=symbols,
		sort={control},
		name={\ensuremath{\vec{e}}},
		description={State prediction error}
	}
	
	\newglossaryentry{error_covariance}{type=symbols,
		sort={control},
		name={\ensuremath{\set{\Sigma}^e}},
		description={State prediction error covariance matrix}
	}
	
	\newglossaryentry{system_dynamics}{type=symbols,
		sort={control},
		name={\ensuremath{\Mat{A},\Mat{B},\Mat{K}}},
		description={Linear vehicle system dynamics}
	}

    \newglossaryentry{tolerance_level}{type=symbols,
		sort={control},
		name={\ensuremath{\kappa}},
		description={Tolerance level, defines the contour line of the bivariate Gaussian distribution}
	}
	
	\newglossaryentry{probability parameter}{type=symbols,
		sort={control},
		name={\ensuremath{\epsilon}},
		description={Probability parameter that defines the level of confidence with which a TV lies within the elliptic region around its predicted state}
	}

	\newglossaryentry{trajectory}{type=symbols,
		sort={control},
		name={\ensuremath{\chi_k^U}},
		description={Trajectory starting at $\gvec{\xi}_k$ and applying $U$}
	}
	
	\newglossaryentry{trajectory_solution}{type=symbols,
		sort={control},
		name={\ensuremath{\chi(k,\gvec{\xi}_0,U)}},
		description={State of a trajectory $\chi_0^U$ at time $k$}
	}
	
	\newglossaryentry{feasible_trajectories}{type=symbols,
		sort={control},
		name={\ensuremath{\Gamma_k}},
		description={Set of feasible trajectories $\chi_k^U$ resulting in $\set{\Xi}_f$}
	}
	
	\newglossaryentry{reachable_set}{type=symbols,
		sort={control},
		name={\ensuremath{\mathcal{R}_k^{\text{TV}}}},
		description={Set of states that can be reached by the TV at time $k$}
	}
	
	\newglossaryentry{set_of_initial_states}{type=symbols,
		sort={control},
		name={\ensuremath{\set{\Xi}^{\text{TV}}_0}},
		description={Set of possible initial TV states}
	}
	
	\newglossaryentry{safety_sets}{type=symbols,
		sort={control},
		name={\ensuremath{\set{\Xi}^{\text{occ}/\text{ex}/\text{safe}}}},
		description={Occupancy, excluded and safe set of states}
	}
	
	\newglossaryentry{inv_set}{type=symbols,
		sort={control},
		name={\ensuremath{\set{\Xi}^{\text{inv}},\set{\Xi}_{\text{f}}}},
		description={Safe control invariant sets with $\set{\Xi}_{\text{f}} \in \set{\Xi}^{\text{safe}}$}
	}
	
    \newglossaryentry{safety_rectangle}{type=symbols,
		sort={control},
		name={\ensuremath{\set{\Xi}^{\text{Rect}}}},
		description={Set of positions describing the SMPC safety rectangle}
	}

    \newglossaryentry{switching_velo}{type=symbols,
		sort={control},
		name={\ensuremath{v_{\text{s}}}},
		description={Switching velocity, defines the maximum engine power}
	}
	
	\newglossaryentry{max_velo_bounds}{type=symbols,
		sort={control},
		name={\ensuremath{v^{\text{up}}_x}},
		description={Maximum longitudinal velocity bound}
	}
	
	\newglossaryentry{velo_bounds}{type=symbols,
		sort={control},
		name={\ensuremath{v^{\text{low/up},A_i}_{x,N}}},
		description={Lower and upper velocity bounds at time $N$ of region $A_i$}
	}
	
	\newglossaryentry{sum_velo_bounds}{type=symbols,
		sort={control},
		name={\ensuremath{v^-, v^+}},
		description={Summarized velocity bounds for all regions $A_i$}
	}
	
    \newglossaryentry{velo_ex}{type=symbols,
		sort={control},
		name={\ensuremath{v^{\text{ex}}_{x,\text{min/max},N}}},
		description={Minimally and maximally possible velocities of the TV at time $N$}
	}
	
	\newglossaryentry{velo_regions}{type=symbols,
		sort={control},
		name={\ensuremath{A_i}},
		description={Region for which the velocity bounds $v^{\text{low/up},A_i}_{x,N}$ hold}
	}

	\newglossaryentry{transition}{type=symbols,
		sort={control},
		name={\ensuremath{\delta}},
		description={Transition width between inner and outer velocity bounds}
	}

	\newglossaryentry{cost}{type=symbols,
		sort={control},
		name={\ensuremath{J}},
		description={Average cost per step}
	}
	
	\newglossaryentry{quad_input_rate}{type=symbols,
		sort={control},
		name={\ensuremath{\mu_{\triangle ^2}, \sigma^2_{\triangle ^2}}},
		description={Mean and variance of the quadratic longitudinal input rate $\triangle u_x^2$}
	}

\newglossaryentry{vector}{type=notation,
	sort={vector},
	name={\ensuremath{\bm{z}_k}},
	description={Vector at timestep $i$}
}
\newglossaryentry{seqi}{type=notation,
	sort={seqi},
	name={\ensuremath{(\cdot)_{(1:k)}}},
	description={Combined variable up to $i$-th step if $(\cdot)^{(i)}$ exists; e.g.: $\bm{z}^{(1:i)}=\begin{bmatrix}\bm{z}^{(1)} & \cdots & \bm{z}^{(i)}\end{bmatrix}^\T$}
}	
\newglossaryentry{seqN}{type=notation,
	sort={seqN},
	name={\ensuremath{(\cdot)}},
	description={Combined variable for \gls{N}/\gls{Nmpc} steps if $(\cdot)^{(i)}$ exists; e.g.: $\bm{z}=\begin{bmatrix}\bm{z}^{(1)} & \cdots & \bm{z}^{(N)}\end{bmatrix}^\T$}
}	
\newglossaryentry{seqNi}{type=notation,
	sort={seqNi},
	name={\ensuremath{(\cdot)_k}},
	description={Combined variable for \gls{Nmpc} steps if $(\cdot)^{(i)}$ exists; e.g.: $\bm{z}_i=\begin{bmatrix}\bm{z}_i^{(i+1)} & \cdots & \bm{z}_i^{(i+N)}\end{bmatrix}^\T$}
}	
\newglossaryentry{matrix}{type=notation,
	sort={matrix},
	name={\ensuremath{Z}},
	description={Matrix}
}		
\newglossaryentry{rv}{type=notation,
	sort={rv},
	name={\ensuremath{\bm{\rv{Z}}_k}},
	description={Random vector at timestep $i$}
}	
\newglossaryentry{set}{type=notation,
	sort={set},
	name={\ensuremath{\set{Z}_k}},
	description={Set of admissable \gls{vector}}
}
\newglossaryentry{I}{type=notation,
	sort={I},
	name={\ensuremath{I_{k\times k}}},
	description={$k$-by-$k$ identity matrix}
}


%% file: chapters/introduction.tex
\section{Introduction}
\label{sec:introduction}

\vspace{-9.5cm}
\mbox{\small This~work~has~been~accepted~to~the~ICRA~2021~workshop~``Opportunities~and~Challenges~with~Autonomous~Racing''.}
\vspace{8.7cm}

Whereas research on automated road vehicles has dominated the past decades, autonomous racing is a field that has only emerged recently. Roborace and the 2021 Indy Autonomous Challenge provide real-world opportunities to apply theoretic results to the racetrack.

Previous work on autonomous racing mainly focuses on lap-time optimization. In \cite{WischnewskiBetzLohmann2019} a model-free learning method is proposed, where the maximum accelerations in longitudinal and lateral direction are adapted. This proposed approach may be incorporated into existing planners and feasibility was demonstrated on a full-size race car used for Roborace. In \cite{StahlEtalLienkamp2019}, trajectories for racing are planned with a multi-layered graph-based planner, verified again on a Roborace vehicle.

Furthermore, learning-based MPC methods have shown promising results for lap-time optimization. In \cite{RosoliaCarvalhoBorrelli2017} an iterative learning MPC framework was proposed that uses information from previous laps, evaluated in simulations. Advances were made in \cite{RosoliaBorrelli2020a}, where the proposed approach was successfully implemented on cars the size of remote control vehicles. A Gaussian process-based learning MPC algorithm is presented in \cite{KabzanEtalZeilinger2019}, improving lap-times while considering safety. The proposed algorithm was applied to a full-size vehicle.

Previous work on autonomous racing majorly focuses on improving lap times; however, a fundamental part of racing is neglected: overtaking other race cars. While Roborace included overtaking maneuvers, these maneuvers were passive ones. Once the trailing vehicle was close enough to the leading vehicle in a specific part of the race track, the leading vehicle had to allow the trailing vehicle to overtake. Active overtaking will be inevitable for participants of the Indy Autonomous Challenge, where active overtaking is required.

In this work, we propose a combined Gaussian Process (GP) and stochastic Model Predictive Control (MPC) method, which allows to actively overtake other race cars.

Gaussian process regression is a powerful non-parametric tool used to infer values of an unknown function given previously collected measurements. In addition to exhibiting very good generalization properties, a major advantage of GPs is that they come equipped with a measure of model uncertainty, making them particularly beneficial for safety-critical applications. These characteristics have made GPs particularly attractive for developing control algorithms \cite{CaponeHirche2019,JagtapEtal2020,CaponeHirche2020,FisacEtal2019}. In the context of autonomous driving, GPs have also seen a rise in interest. In \cite{HewingEtal2018,KabzanEtalZeilinger2019}, GP regression is used to improve the model of the autonomous vehicle using collected data, which in turn leads to an improvement in control performance. GPs have also been employed to predict the behavior of cut-in maneuvers of surrounding vehicles and obtain safe autonomous vehicle control \cite{YoonEtal2021}.

Stochastic MPC (SMPC) has mostly been studied for road vehicles. SMPC allows to treat constraints in a probabilistic way, enabling less conservative solutions \cite{Mesbah2016, FarinaGiulioniScattolini2016}. In \cite{CarvalhoEtalBorrelli2014} an SMPC trajectory planner for automated vehicles is presented, considering the most likely future maneuver of surrounding vehicles and Gaussian prediction uncertainty. The Gaussian uncertainty allows to analytically reformulate the probabilistic constraint into a deterministic formulation that is tractable by a solver. In \cite{SchildbachBorrelli2015} a sampling-based SMPC approach for automated driving is proposed, based on scenario MPC \cite{SchildbachEtalMorari2014}. The approach of \cite{SchildbachBorrelli2015} is extended in \cite{CesariEtalBorrelli2017} to account for more complex vehicle behavior. Whereas standard SMPC approaches allow a small probability of collision, a safe SMPC framework is developed in \cite{BruedigamEtalLeibold2021b}.

This paper outlines a combined GP and SMPC approach for autonomous overtaking maneuvers in racing. The major challenge is to plan trajectories for a controlled race car such that a leading race car may be passed. Based on previous behavior of the leading vehicle, given the interaction between both vehicles, a GP is trained. The GP predictions for the leading vehicle are then used in an SMPC algorithm to plan efficient overtaking maneuvers. Whereas we only show preliminary results here, the proposed method has the potential to be a powerful method in autonomous racing. Ideally, the GP identifies weaknesses in the driving behavior of the leading vehicle while the controlled trailing vehicle is trying to overtake. The SMPC planner allows to efficiently consider the GP output and to plan optimistic vehicle trajectories, which are fundamental for racing. Given an increased sample set of data, the controlled race car increases its chances of finding the right spot on a race track and a suitable driving approach to successfully overtake. 

This paper is structured as follows. Section~\ref{sec:problem} introduces the vehicle model considered for the prediction. In Section~\ref{sec:method} the proposed GP-SMPC method for race overtaking maneuvers is described. Simulation results are given in Section~\ref{sec:results}, while a conclusion and outlook follows in Section~\ref{sec:conclusion}.

%% file: chapters/problem.tex
\section{Vehicle Model}
\label{sec:problem}

We consider two vehicles. The controlled race car is denoted as the ego vehicle (EV), whereas the race car to be overtaken is a target vehicle (TV). 

MPC requires prediction models for both vehicles. The TV prediction, used by the EV, is based on the GP described in Section~\ref{sec:traj_gen}, whereas the actual TV behavior is described in Section~\ref{sec:simu_setup} of the simulation part. For the EV, a kinematic bicycle model is used, given by the continuous-time system
\begin{IEEEeqnarray}{rl}
\IEEEyesnumber \label{eq:bicyclemodel}
\dot{s} &= v \cos(\phi + \alpha), \IEEEyessubnumber \label{eq:bm_s}\\
\dot{d} &= v \sin(\phi + \alpha), \IEEEyessubnumber \label{eq:bm_d} \\
\dot{\phi} &= \frac{v}{l_\text{r}}\sin{\alpha}, \IEEEyessubnumber \label{eq:bm_phi} \\
\dot{v} &= a, \IEEEyessubnumber \label{eq:bm_a} \\
\alpha &= \arctan \lr{\frac{l_\text{r}}{l_\text{r}+l_\text{f}} \tan \delta}, \IEEEyessubnumber \label{eq:bm_alpha}
\end{IEEEeqnarray}
where $l_\text{r}$ and $l_\text{f}$ represent the distances from the vehicle center of gravity to the rear and front axles, respectively. The state and input vectors are $\bm{\xi} = [s,d,\phi,v]^\top$ and $\bm{u} = [a,\delta]^\top$, respectively, with vehicle velocity $v$, acceleration $a$, and steering angle $\delta$. The longitudinal position along the road is $s$, the lateral vehicle deviation from the race track centerline is $d$, and the orientation of the vehicle with respect to the road is $\phi$. We summarize the nonlinear vehicle model \eqref{eq:bicyclemodel} as $\dot{\bm{\xi}} = \bm{f}^\text{c}\lr{\bm{\xi},\bm{u}}$.

Efficient MPC requires a discrete-time prediction model, which is obtained by first linearizing \eqref{eq:bicyclemodel} at the current EV state $\bm{\xi}^* = \bm{\xi}_0$ and the EV input $\bm{u}^* = [0,0]^\top$ and then discretizing with sampling time $T$. These steps yield the time-discrete system
\begin{IEEEeqnarray}{rl}
\IEEEyesnumber \label{eq:d_evsys}
\bm{\xi}_{k+1} &= \bm{\xi}_0 + T \bm{f}^\text{c} \lr{\bm{\xi}_0, \bm{0}} + \bm{A}_\text{d} \lr{\bm{\xi}_k - \bm{\xi}_0}  + \bm{B}_\text{d} \bm{u}_k \IEEEyessubnumber \IEEEeqnarraynumspace \\ 
&= \bm{f}^\text{d}\lr{\bm{\xi}_0, \bm{\xi}_k, \bm{u}_k}  \IEEEyessubnumber 
\end{IEEEeqnarray}
with the discretized system matrices $\bm{A}_\text{d}$, $\bm{B}_\text{d}$ and a nonlinear term $\bm{f}^\text{c} \lr{\bm{\xi}_0, \bm{0}}$. Details on the linearization and discretization are provided in \cite{BruedigamEtalLeibold2021b, BruedigamEtalLeibold2020d}.

The EV is subject to constraints. We consider input constraints 
\begin{IEEEeqnarray}{rrl}
\IEEEyesnumber \label{eq:constraints_input}
\bm{u}_\text{min} \leq & \gls{uk} &\leq \bm{u}_\text{max} \IEEEyessubnumber \label{eq:con_u}\\
\Delta\bm{u}_\text{min} \leq & \Delta\gls{uk} &\leq \Delta\bm{u}_\text{max} \IEEEyessubnumber \label{eq:con_du}
\end{IEEEeqnarray}
limiting the absolute value and rate of change of the acceleration and steering angle, where $\Delta \bm{u}_{k+1} = \bm{u}_{k+1} - \gls{uk}$. In addition, the road constraint is given by $d_\text{min} \leq \gls{dk} \leq d_\text{max}$ and we require a non-negative velocity $\gls{vk} \geq 0$. In the following, the input and state constraints are summarized by the set of admissible inputs \gls{UU} and the set of admissible states \gls{XX}. Additional collision avoidance constraints will be designed in the following section.

%% file: chapters/method.tex
\section{GP-based SMPC for Autonomous Racing}
\label{sec:method}

Autonomous racing requires overtaking maneuvers. In order to plan successful overtaking maneuvers, the EV needs a precise prediction of the future behavior of the TV. Here, we propose a combined GP and SMPC framework, where GP is used to predict the future TV behavior and SMPC plans optimistic EV trajectories, facilitating an overtaking maneuver to pass the TV.

In the following, we first present details on the GP design. Then, the generation of safety constraints is briefly addressed and, depending on the GP output, these constraints are tightened. Eventually, the tightened safety constraints are included into an SMPC optimal control problem to avoid collisions. 

\subsection{Gaussian Processes}
\label{sec:gp_det}

Gaussian processes are used to infer the values of an unknown function given measurement data $\mathcal{D} =\left\{\bm{x}_n, \bm{y}_n\right\}_{n=1}^{N}$, where the training inputs
\begin{align}
    \begin{split}
        \bm{x}_n \coloneqq \left(\bm{\xi}_n^{\top}, (\bm{\xi}^\text{TV}_n)^{\top} \right)^{\top}
    \end{split}
\end{align}
correspond to the concatenation of the EV and TV states, and the training outputs
\begin{align}
    \begin{split}
    \bm{y}_n \coloneqq \left({\bm{\xi}}^\text{TV}_{n+1}-\bm{\xi}^\text{TV}_{n}\right)^{\top}
    \end{split}
\end{align}
are the difference between the TV states for two time steps.

A Gaussian process is formally defined as a collection of random variables, any subset of which is jointly normally distributed \cite{rasmussen2003gaussian}. It is fully specified by a prior mean, which we set to zero without loss of generality, and a kernel function $\kappa: \mathbb{R} \times \mathbb{R} \rightarrow \mathbb{R}$. The kernel $\kappa(\cdot,\cdot)$ encodes function properties and any prior assumptions, e.g., Lipschitz continuity, periodicity and magnitude. In the following, we employ a squared-exponential kernel 
\begin{align}
    \kappa( \bm{x},  \bm{x}') = \sigma^2 \exp\left(-\frac{(\bm{x}-\bm{x}')^{\top}\bm{L}^{-2}(\bm{x}-\bm{x}')}{2}\right),
\end{align}
which can approximate continuous functions in compact spaces arbitrarily accurately \cite{MicchelliEtal2006}.

By modeling the state transition dynamics ${\bm{\xi}}^\text{TV}_{n+1}-\bm{\xi}^\text{TV}_{n}$ of the TV with a GP, we implicitly assume that any set of evaluations is jointly normally distributed. By conditioning the GP on the training data $\mathcal{D}$, we obtain the posterior mean and variance for the $d$-th entry of the transition dynamics at an arbitrary point $\bm{x}^*$,
\begin{align}
\label{eq:posteriordist}
    \begin{split}
        \mu_d(\bm{x}^* \vert \mathcal{D}) &= \bm{\kappa}^{\top} \bm{K}^{-1}\bm{\gamma}_{d} \\
        \sigma_d(\bm{x} \vert \mathcal{D}) & = \kappa^* - \bm{\kappa}^{\top} \bm{K}^{-1} \bm{\kappa},
    \end{split}
\end{align}
where $\kappa^*=\kappa(\bm{x}^*,\bm{x}^*)$, $\bm{\kappa} = (\kappa(\bm{x}_1, \bm{x}^*),\ldots,\kappa(\bm{x}_n, \bm{x}^*))$, the entries of the matrix $\bm{K}$ are given by $K_{ij} = \kappa(\bm{x}_i,\bm{x}_j)$, and $\bm{\gamma}_d =(y_{1,d},\ldots, y_{n,d} )$ concatenates the training outputs corresponding to the $d$-th entry.

\subsection{Generating Sample TV Trajectories}
\label{sec:traj_gen}
We employ the GP model described in \Cref{sec:gp_det} to generate $M$ sample TV trajectories
\begin{align}
    \begin{split}
        \bm{\xi}^{\text{TV},(m)}_k, ~~~~k \in \{1,\dots,N\},~~~~m \in \{1,\dots,M\}.
    \end{split}
\end{align}
To this end, we sequentially draw a sample from the posterior GP distribution \eqref{eq:posteriordist}, apply the sampled dynamics to the TV prediction, and then condition the GP on the sampled point, similarly to \cite{CaponeHirche2020}.

The GP sample trajectories correspond to a computational complexity of order $\mathcal{O}(MN^3)$, which can become cumbersome for long horizons. However, the scalability of GPs can be improved considerably by employing several different approximations, e.g., by employing a set of inducing points or approximating the squared-exponential kernel by a finite-dimensional feature map \cite{WilsonEtal2020}.

From the sample trajectories, we deduce the mean and variance of the TV states at the prediction time steps
\begin{align}
    \begin{split}
        \overline{\bm{\xi}}^{\text{TV}}_k &= \frac{1}{M}\sum_{m=1}^M{\bm{\xi}}^{\text{TV},(m)}_k, ~~~~k \in \{1,\dots,N\}, \\
        \bm{\Sigma}^2_{k} &= \frac{1}{M-1}\sum_{m=1}^M \left({\bm{\xi}}^{\text{TV},(m)}_k - \overline{\bm{\xi}}^{\text{TV}}_k \right)^{\top}\left({\bm{\xi}}^{\text{TV},(m)}_k - \overline{\bm{\xi}}^{\text{TV}}_k \right),
    \end{split}
\end{align}
which are later used for the SMPC part. The diagonal elements of $\bm{\Sigma}^2_{k}$ are $[\Sigma^2_{k,x}, \Sigma^2_{k,v_x}, \Sigma^2_{k,y}, \Sigma^2_{k,v_y}]$.

\subsection{Constraint Generation}
\label{sec:con_gen}

We generate constraints similar to \cite{BruedigamEtalLeibold2021b}. A safety rectangle with length 
\begin{IEEEeqnarray}{c}
a_\text{r} = l_\text{veh} + \tilde{a}_\text{r}\lr{\bm{\xi}, \bm{\xi}^\text{TV}}.
\end{IEEEeqnarray}
and width
\begin{IEEEeqnarray}{c}
b_\text{r} = w_\text{veh} + \varepsilon_\text{safe}
\end{IEEEeqnarray}
is designed that surrounds the TV, where $l_\text{veh}$ and $w_\text{veh}$ denote the vehicle length and width, respectively. The size of the safety rectangle length depends on the velocity difference between the EV and TV, summarized by the term $\tilde{a}_\text{r}\lr{\bm{\xi}, \bm{\xi}^\text{TV}}$. A lateral safety distance parameter $\varepsilon_\text{safe}$ is used. Details are given in \cite{BruedigamEtalLeibold2021b, BruedigamEtalLeibold2020d}.

\begin{figure}
\centering
\includegraphics[width = \columnwidth]{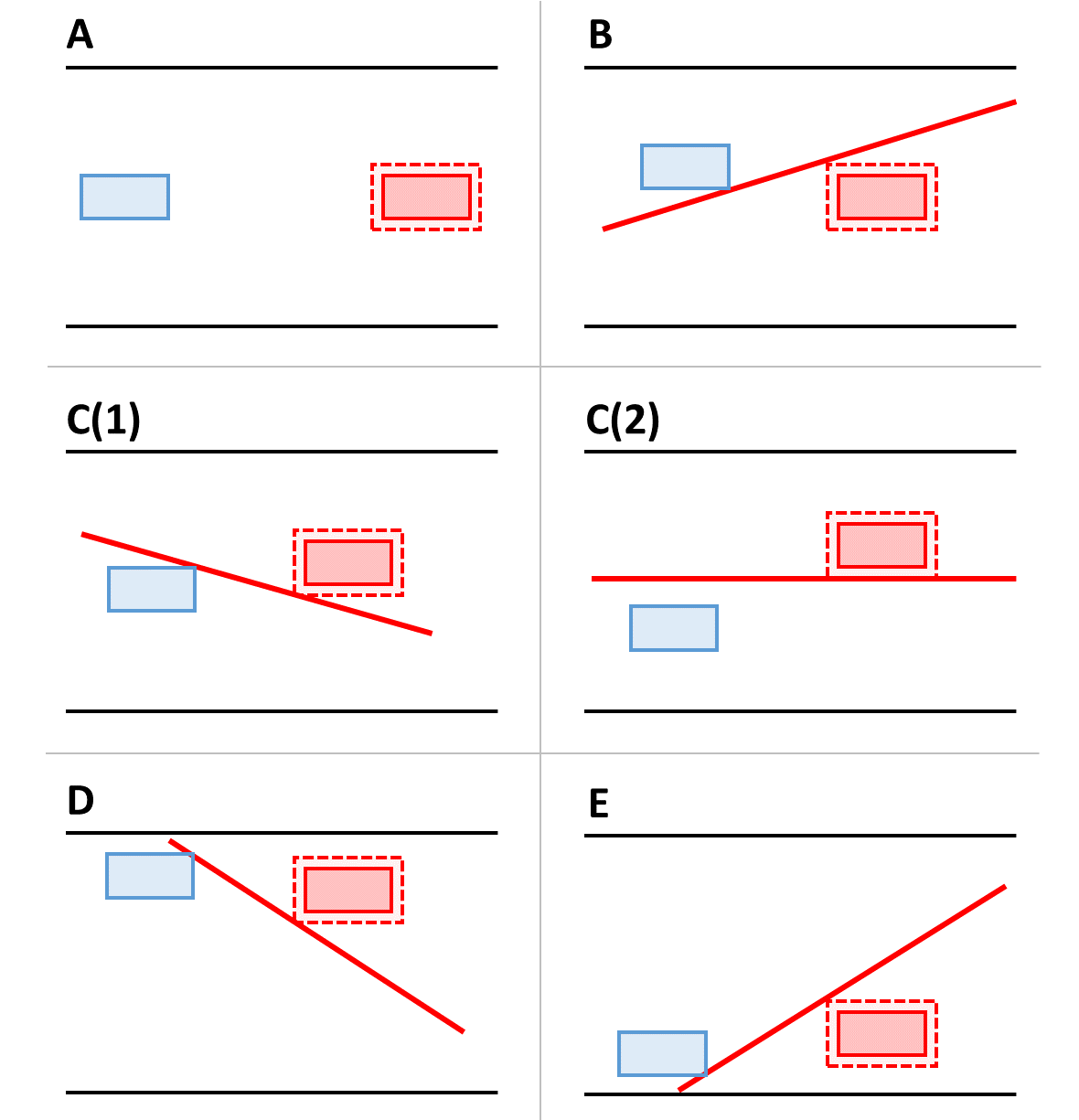}
\caption{Constraint cases. Driving direction is from left to right. The \gls{ev} and \gls{tv} are shown in blue and red, respectively. The dashed red line represents the safety area around the \gls{tv}.}
\label{fig:constraintcases}
\end{figure}

Depending on the positioning of the EV with respect to the TV, different cases are considered for the constraint generation. The cases are summarized in Table~\ref{tab:constraint_cases} 
\begin{table}
\centering
\caption{Constraint Generation Cases}
\label{tab:constraint_cases}
\begin{tabular}{l    l     l  }
\toprule
case & EV setting (w.r.t. TV) & constraint\\
\midrule
A & large distance  & no constraint  \\
\addlinespace
B & left of TV & inclined (pos.) constraint\\
\addlinespace
C & right of TV & inclined (neg.) constraint\\
\addlinespace
D & \begin{tabular}{@{}l@{}} left of TV\\ (close to road limit) \end{tabular}  & inclined (neg.) constraint\\
\addlinespace
E & \begin{tabular}{@{}l@{}} right of TV\\ (close to road limit) \end{tabular} & inclined (pos.) constraint\\
\bottomrule
\end{tabular}
\end{table}
and illustrated in Fig~\ref{fig:constraintcases}. 
Inclined constraints are limited to a horizontal level, as shown in case C(2) in Fig.~\ref{fig:constraintcases}. Once the EV gets too close to the road boundary, i.e., overtaking is impossible, the constraint is chosen such that the EV plans to overtake on the other side (cases D and E).

Based on this case differentiation, linear safety constraints follow, given by 
\begin{IEEEeqnarray}{c}
0 \geq q_y\lr{\bm{\xi}_0, \gls{xitv}} d_k + q_x\lr{\bm{\xi}_0, \gls{xitv}} s_k + q_t\lr{\bm{\xi}_0, \gls{xitv}} \label{eq:c_smpc}
\end{IEEEeqnarray}
with the coefficients $q_y$ and $q_x$ for the \gls{ev} states $d_k$ and $s_k$, and the intercept $q_t$. The coefficients $q_y$, $q_x$, and $q_t$ depend on the current \gls{ev} state $\bm{\xi}_0$ and the predicted \gls{tv} states \gls{xitv}. Whereas future TV state predictions are used, only the current \gls{ev} state is considered in order to generate linear constraints.

\subsection{SMPC Details}
\label{sec:smpc_det}

In the previous computation of the safety rectangle, no TV prediction uncertainty was considered. In the following, we extend the safety rectangle to account for TV prediction uncertainty, yielding the updated safety rectangle length and width values
\begin{IEEEeqnarray}{rl}
\IEEEyesnumber \label{eq:ab_rectangle_cc}
b_{\text{r},k} &= w_\text{veh} + e_{x,k,\eta} \IEEEyessubnumber \\
a_{\text{r},k} &= l_\text{veh} + \tilde{a}_\text{r}\lr{\bm{\xi}_0, \gls{xitv}} + e_{y,k,\eta} \IEEEyessubnumber
\end{IEEEeqnarray}
with constraint tightening according to $e_{x,k,\eta}$ and $e_{y,k,\eta}$ as discussed next.

SMPC considers probabilistic constraints, i.e., chance constraints, of the form
\begin{IEEEeqnarray}{c}
\text{Pr}\left(\bm{\xi}_k \in \bm{\Xi}_{k,\text{safe}} \right) \geq \beta, \label{eq:cc_ev}
\end{IEEEeqnarray}
which must be fulfilled with a probability larger than the risk parameter $\beta$. The safe set $\bm{\Xi}_{k,\text{safe}}$ depends on the TV safety rectangle. It is not possible to directly solve \eqref{eq:cc_ev}; hence, a reformulation is necessary.

Based on the approximated error covariance matrices $\bm{\Sigma}^2_k$, considering only diagonal elements, the TV safety rectangle adaptations $e_{x,k,\eta}$ and $e_{y,k,\eta}$ are obtained by
\begin{IEEEeqnarray}{rl}
\IEEEyesnumber \label{eq:ellipse_axes}
e_{x,k,\eta}= \Sigma_{k,x} \sqrt{\eta}  \IEEEyessubnumber \\
e_{y,k,\eta}= \Sigma_{k,y} \sqrt{\eta}. \IEEEyessubnumber
\end{IEEEeqnarray}
with the \textit{chi-squared distribution}
\begin{IEEEeqnarray}{c}
\eta = \chi^2_2 (1-\beta)
\end{IEEEeqnarray}
as derived in detail in \cite{BruedigamEtalLeibold2021b}.

The generated safety constraints based on the adapted safety rectangles are now included into an SMPC optimal control problem.

\subsection{Optimal Control Problem}
\label{sec:ocp}

Given the constraint generation and constraint tightening, we formulate the deterministic representation of the SMPC optimal control problem
\begin{IEEEeqnarray}{rl}
\IEEEyesnumber \label{eq:smpc}
	V^* &= \min_{\bm{U}} \sum_{k=1}^{N} \norm{\Delta \bm{\xi}_k}_{\bm{Q}} + \norm{\bm{u}_{k-1}}_{\bm{R}} + \norm{\Delta\bm{u}_{k-1}}_{\bm{S}}
	 \IEEEyessubnumber \IEEEeqnarraynumspace \\
	\text{s.t. } & \bm{\xi}_{k+1} = \bm{f}^\text{d}\lr{\bm{\xi}_0, \bm{\xi}_k, \bm{u}_k} \IEEEyessubnumber \IEEEeqnarraynumspace \label{eq:smpc_dynamics}\\
	& \bm{\xi}_k \in \gls{XX} ~~~~\forall k \in \{1,\dots,N\} , \IEEEyessubnumber\\
	& \bm{u}_k \in \mathcal{U} ~~~\forall k \in \{0,\dots,N-1\} ,\IEEEyessubnumber \IEEEeqnarraynumspace\\
	& 0 \geq q_y\lr{\bm{\xi}_0, \gls{xitv}} y_k + q_x\lr{\bm{\xi}_0, \gls{xitv}}  x_k + q_t\lr{\bm{\xi}_0, \gls{xitv}} \IEEEeqnarraynumspace \nonumber \\
	&~~\forall k \in \{0,\dots,N\} \IEEEyessubnumber  \label{eq:smpc_cc}
\end{IEEEeqnarray}
where $\norm{\bm{z}}_{\bm{Z}} = \bm{z}^\top \bm{Z} \bm{z}$ and $\Delta \bm{\xi}_k = \bm{\xi}_k - \bm{\xi}_{k, \text{ref}}$ with the \gls{ev} reference state $\bm{\xi}_{k, \text{ref}}$. The weighting matrices are given by $\bm{Q}$, $\bm{S}$, and $\bm{R}$.

The resulting optimal control problem is a quadratic program and can be solved efficiently. The constraint tightening steps are performed before the optimal control problem is solved. 

%% file: chapters/results.tex
\section{Simulation Results}
\label{sec:results}

In the following, we briefly analyze the proposed SMPC approach in a simulation scenario. 

\subsection{Simulation Setup}
\label{sec:simu_setup}

\begin{figure}
\centering
\includegraphics[width = \columnwidth]{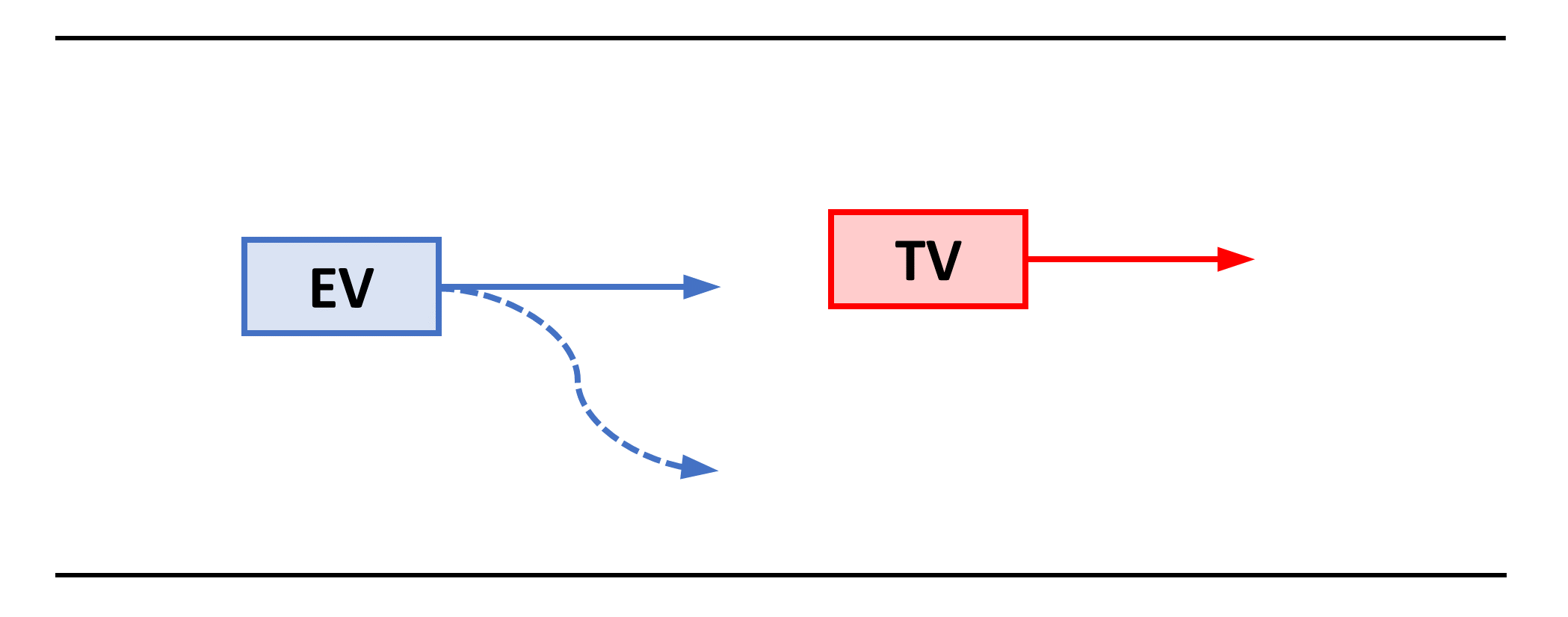}
\caption{Racing scenario with two possible EV trajectories.}
\label{fig:scenario_setup}
\end{figure}

We consider a simple race scenario, where the EV intends to overtake the TV on a straight road, as illustrated in Fig.~\ref{fig:scenario_setup}. 

All units are given as SI units. The simulations are carried out in Matlab. The MPC algorithm is based on the NMPC toolbox \cite{GruenePannek2017}, using the \textit{fmincon} solver.

The MPC algorithm uses a sampling time $T=0.2$ and a horizon $N=10$. The linearized, discretized EV model \eqref{eq:d_evsys} is used for the prediction. The optimized MPC inputs are then applied to the nonlinear, time-continuous EV model \eqref{eq:bicyclemodel} in the simulation. Important simulation parameters are summarized in Table~\ref{tab:sim_params}. 
\begin{table}
\centering
\caption{General Simulation Parameters}
\label{tab:sim_params}
\begin{tabular}{l    l     l }
\toprule
scalars &  vectors & matrices \\
\midrule
$w_\text{lane} = 12$ & $\bm{u}_\text{max} = [10, 0.2]^\top$ & $\bm{Q} = \text{diag}(0, 0.25, 0.2, 10)$ \\
\addlinespace
$l_\text{veh} = 5$ & $\bm{u}_\text{min} = [\text{-}15, \text{-}0.2]^\top$ & $\bm{R} = \text{diag}(0.33, 5)$ \\
\addlinespace
$w_\text{veh} = 2$ & $\bm{u}^{\text{TV}}_\text{max} = [10, 0.4]^\top$ 
& $\bm{S} = \text{diag}(0.33, 15)$ \\
\addlinespace
$l_\text{f} = l_\text{r} = 2$ & $\bm{u}^{\text{TV}}_\text{min} = [\text{-}15, \text{-}0.4]^\top$ 
&  \\
\addlinespace
$\varepsilon_\text{safe} = 0.5$ & & \\
\bottomrule
\end{tabular}
\end{table}
In case of an infeasible SMPC optimal control problem, the previously optimized solution is applied.

A simple TV motion planner was implemented, depending on previous and current vehicle states. The TV states are $\bm{\xi}^\text{TV} = [x^\text{TV}, v_x^\text{TV}, y^\text{TV}, v_y^\text{TV}]^\top$ with longitudinal and lateral inputs $u_x^\text{TV}$ and $u_y^\text{TV}$. The TV state update is computed by
\begin{IEEEeqnarray}{c}
\bm{\xi}^{\text{TV}}_{k+1} = \bm{A} \gls{xitv}+\bm{B} \gls{utv}  \label{eq:tvsys}
\end{IEEEeqnarray}
with
\begin{IEEEeqnarray}{c}
\bm{A} = \begin{bmatrix} 1 & \gls{dt} & 0 & 0  \\
											0 & 1 & 0 & 0 \\
											0 & 0 & 1 & \gls{dt} \\
											0 & 0 & 0 & 1 \end{bmatrix},
	~\bm{B} = \begin{bmatrix} 0.5 \gls{dt}^2 & 0 \\
											\gls{dt} & 0 \\ 0 & 0.5 \gls{dt}^2 \\ 0 & \gls{dt} \end{bmatrix}
  \label{eq:tv_AB}
\end{IEEEeqnarray}
and
\begin{IEEEeqnarray}{rl}
\IEEEyesnumber
\gls{utv} &= \bm{K}^\text{TV} \lr{\gls{xitv} - \gls{xitvref}}, \IEEEyessubnumber\label{eq:tv_feedback}\\
\bm{K}^\text{TV} &= \begin{bmatrix} 0 & -0.55 & 0 & 0 \\ 0 & 0 & -0.63 & -1.15 \end{bmatrix} \IEEEyessubnumber
\end{IEEEeqnarray}
based on a TV reference state \gls{xitvref} and a stabilizing feedback matrix $\bm{K}^\text{TV}$. The TV inputs are limited by $\bm{u}^{\text{TV}}_\text{max}$ and $\bm{u}^{\text{TV}}_\text{min}$. 

The TV reference state is then adapted depending on the EV and TV setting. Once the EV has passed the TV, the TV aims to drive straight. As long as the EV is located behind the TV, the TV tries to block the EV from overtaking. This is done by setting the lateral TV position reference to the current EV lateral position, while maintaining a constant longitudinal velocity. Similar to many real-world races, the TV is only allowed to choose a maneuver once, i.e., the TV cannot continuously drive left and right to block the EV but must only move into one direction or drive straight.

\subsection{Overtaking Maneuver}
\label{sec:ot}

We analyze the proposed method in a simple race scenario. The initial EV state is $\bm{\xi}_0 = [0, 0, 0, 60]^\top$ and the initial TV state is $\bm{\xi}^{\text{TV}}_0 = [80, 50, -2.5, 0]^\top$. The EV aims at maintaining its initial velocity, its center road position, and a straight orientation with respect to the road, i.e., $v_\text{ref} = 60$, $d_\text{ref} = 0$, and $\phi_\text{ref} = 0$, respectively.

\begin{figure}
\centering
\includegraphics[width = \columnwidth]{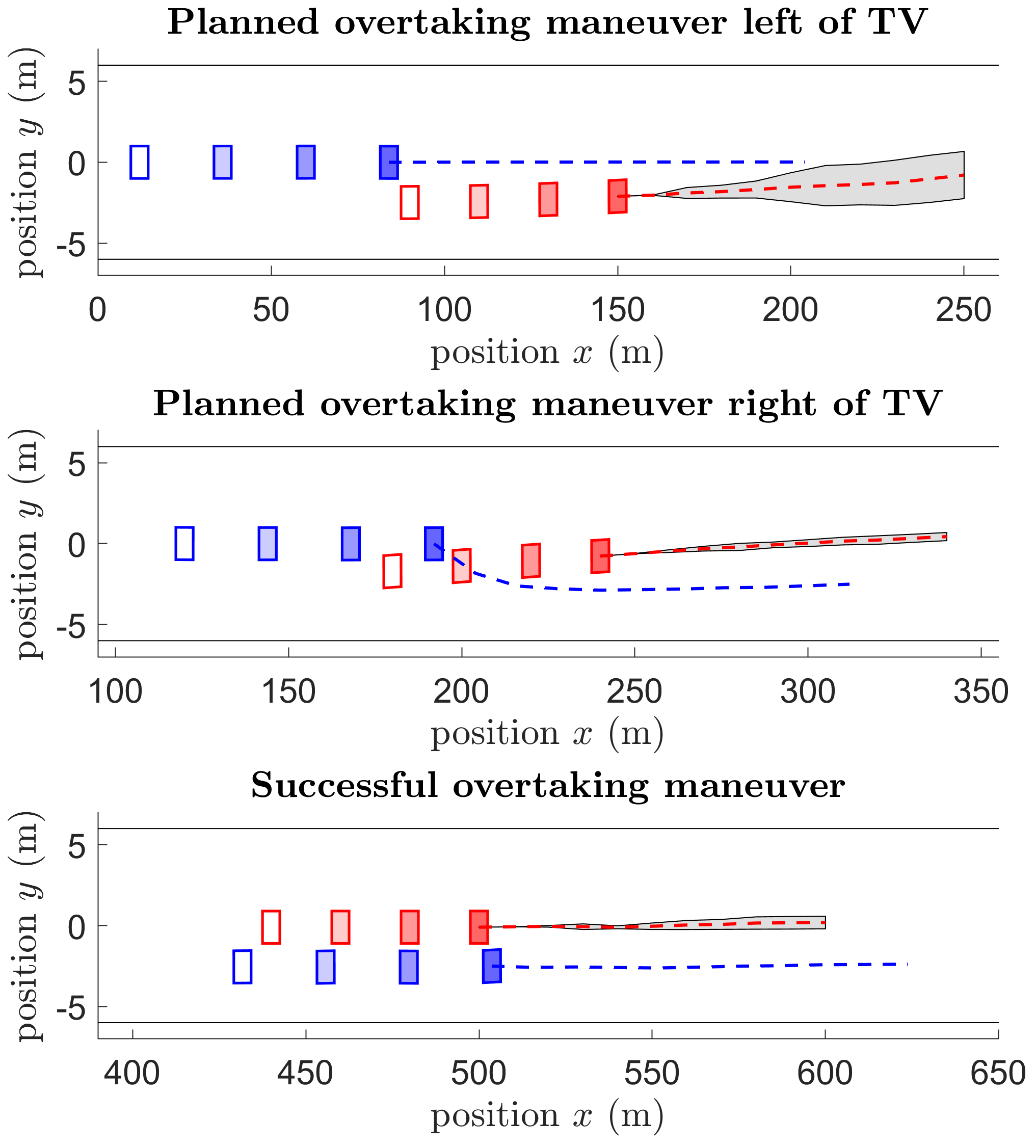}
\caption{Shots of the overtaking maneuver. Fading boxes show past vehicle states. The EV is shown in blue, the TV in red. Predicted EV and TV trajectories are indicated by dashed lines. The GP $2\sigma$ confidence region for the lateral TV motion is visualized by the gray area.}
\label{fig:shots}
\end{figure}

\begin{figure}
\centering
\includegraphics[width = \columnwidth]{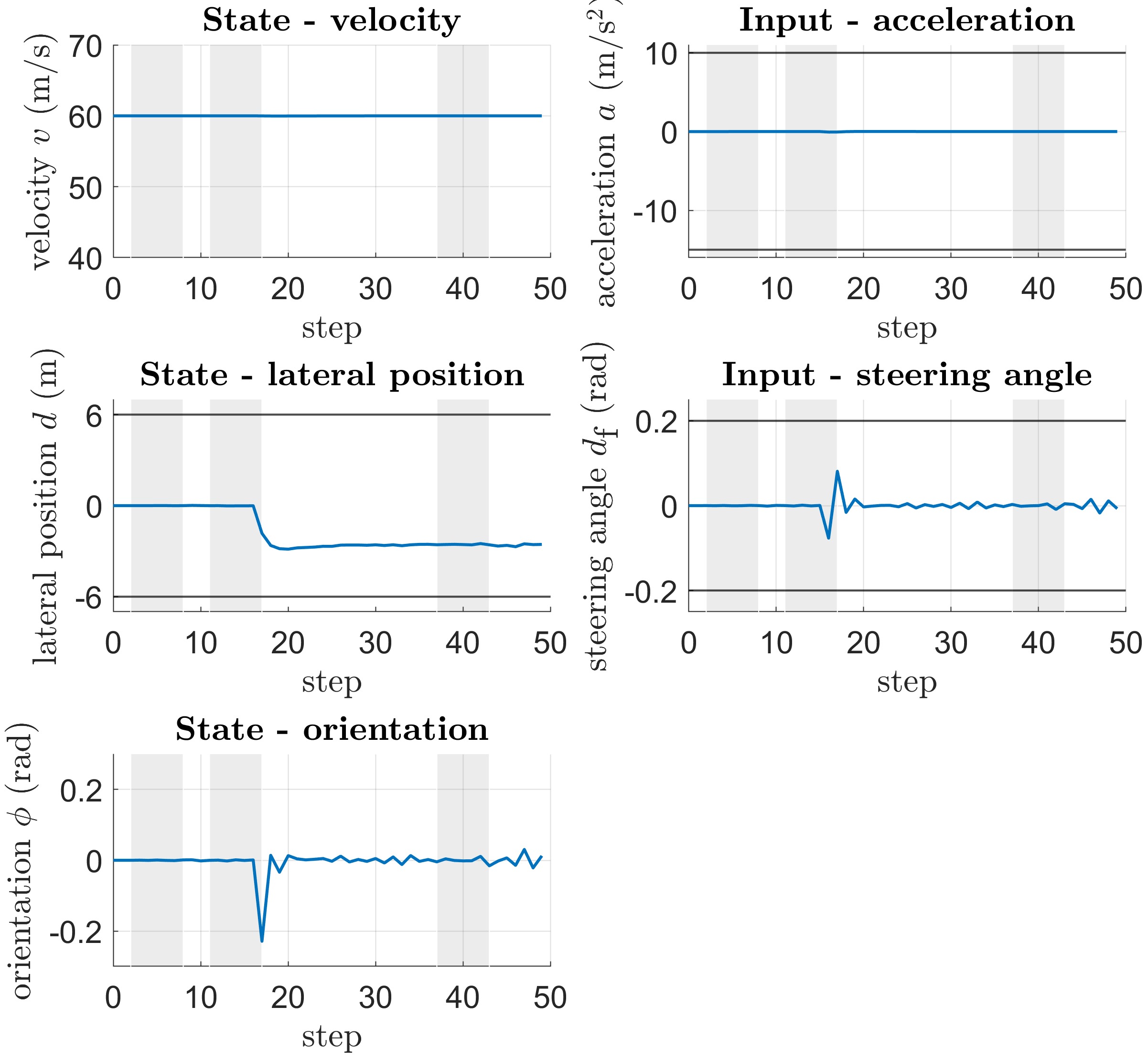}
\caption{EV states and inputs. Gray areas are displayed in Fig.~\ref{fig:shots}.}
\label{fig:states}
\end{figure}

Shots of the vehicle configuration are shown in Fig.~\ref{fig:shots}. 
The EV states and inputs are displayed in Fig.~\ref{fig:states}. 

Initially, the EV attempts to overtake the TV on the left. In the beginning, the TV moves towards the left to block the EV overtaking maneuver on the left. At step $17$, the GP prediction indicates that the TV will continue its move to the left. As more data is available, the GP variances decrease. From this point on, it is more beneficial for the EV to switch its strategy and attempt to overtake the TV on the right. As the TV is only allowed to change its lateral direction once, the EV successfully overtakes the TV eventually. Throughout the simulation, the EV maintains its 

%% file: chapters/discussion.tex

%% file: chapters/conclusion.tex
\section{Conclusion}
\label{sec:conclusion}

In this work, we proposed a method where the results of Gaussian process learning are used by stochastic MPC to plan overtaking maneuvers in autonomous racing. The presented work presents a starting point regarding GP and SMPC based research for overtaking in autonomous racing. 

Obvious extensions include using a dynamic vehicle model for the MPC prediction and a more sophisticated Gaussian process learning approach. The next step would then be to extend the method such that a full lap and race with multiple surrounding vehicles is possible. 

Eventually, the aim is to develop a method for competitive racing: First, the controlled autonomous race car learns weak spots of the leading vehicle while driving behind the leading vehicle. Then, the stochastic MPC approach allows to optimistically overtake the leading vehicle at the right part of the race track, based on the results of the Gaussian process.